\pdfoutput=1

\documentclass[11pt]{article}
\usepackage{acl}

\usepackage{times}
\usepackage{latexsym}

\usepackage[T1]{fontenc}

\usepackage[utf8]{inputenc}

\usepackage{microtype}

\usepackage{inconsolata}

\usepackage{graphicx}
\usepackage{amsmath}
\usepackage{amssymb}
\usepackage[shortlabels]{enumitem}
\usepackage[font={small}]{caption}
\usepackage[multiple]{footmisc}
\usepackage{multirow}
\usepackage{booktabs}
\usepackage{mathabx}  
\usepackage{amsthm}   
\usepackage{hyperref}       
\usepackage{url}            
\usepackage{xspace}

\usepackage{algpseudocode}
\usepackage{pifont}

\newcommand\comment[1]{}

\def\model{\textsc{DOTA}\xspace}

\title{Exploring the limits of decoder-only models trained on \\public speech recognition corpora}

\author{Ankit Gupta\thanks{ankitgupta.iitkanpur@gmail.com, gsaon@us.ibm.com, bedk@us.ibm.com} \qquad\quad George Saon \qquad\quad Brian Kingsbury
\\\\ IBM Research}

\begin{document}
\maketitle

\begin{abstract}
The emergence of industrial-scale speech recognition (ASR) models such as Whisper and USM, trained on 1M hours of weakly labelled and 12M hours of audio only proprietary data respectively, has led to a stronger need for large scale public ASR corpora and competitive open source pipelines. Unlike the said models, large language models are typically based on Transformer decoders, and it remains unclear if decoder-only models trained on public data alone can deliver competitive performance.
In this work, we investigate factors such as choice of training datasets and modeling components necessary for obtaining the best performance using public English ASR corpora alone. Our \textbf{D}ecoder-\textbf{O}nly \textbf{T}ransformer for \textbf{A}SR (DOTA) model comprehensively outperforms the encoder-decoder open source replication of Whisper (OWSM) on nearly all English ASR benchmarks and outperforms Whisper large-v3 on 7 out of 15 test sets. We release our codebase and model checkpoints under permissive license.
\end{abstract}

\section{Introduction}\label{sec:intro}

Attention-based models \cite{vaswani2017attention} have been successful across many areas of machine learning \cite{Jumper2021HighlyAP,videopoet}. In particular, large language models (LLMs) comprising decoder-only Transformers pre-trained on large amounts of unlabelled text have become the standard in natural language processing, exhibiting impressive amounts of linguistic and world knowledge \cite{mixtral}. 

In contrast to LLMs, the best performing ASR models are typically based on Conformers \cite{conformer} trained with a
connectionist temporal classification \cite{ctc} or RNN transducer \cite{rnn-t} objective. This method has been highly successful and is employed in nearly all the best performing ASR models \cite{nemo} such as USM \cite{usm} which also serves as the speech encoder of Gemini v1 \cite{geminiv1}. However, as these methods use a monotonic inductive bias specific to ASR, it is natural to investigate if Transformers trained to autoregressively generate text can deliver competitive performance. This was answered with the introduction of Whisper, a Transformer encoder-decoder model that is highly competitive with the best performing CTC-Conformer and RNN-T-Conformer pipelines on several speech recognition and translation benchmarks across multiple languages \cite{whisper}.

Whisper (large-v3) is trained on 1M hours of proprietary speech-text data and 4M hours of pseudo-labelled audio, so it remains to be answered whether similar performance could be achieved via public ASR data alone. Unlike Whisper, which is an encoder-decoder model, most LLMs are decoder-only, and it is natural to investigate the performance of decoder-only models. The OWSM model \cite{owsm} is a step towards this direction and comprises a Whisper-style encoder-decoder model trained on a compilation of public multilingual ASR corpora. However, as OWSM models are encoder-decoder Transformers trained using an additional CTC-based loss, it remains to be answered if conventional Transformer decoder training, similar to that for LLMs \cite{Brown2020LanguageMA}, suffices for competitive performance.

\begin{figure*}[t]
  \centering  
  \includegraphics[width=1.0\textwidth]{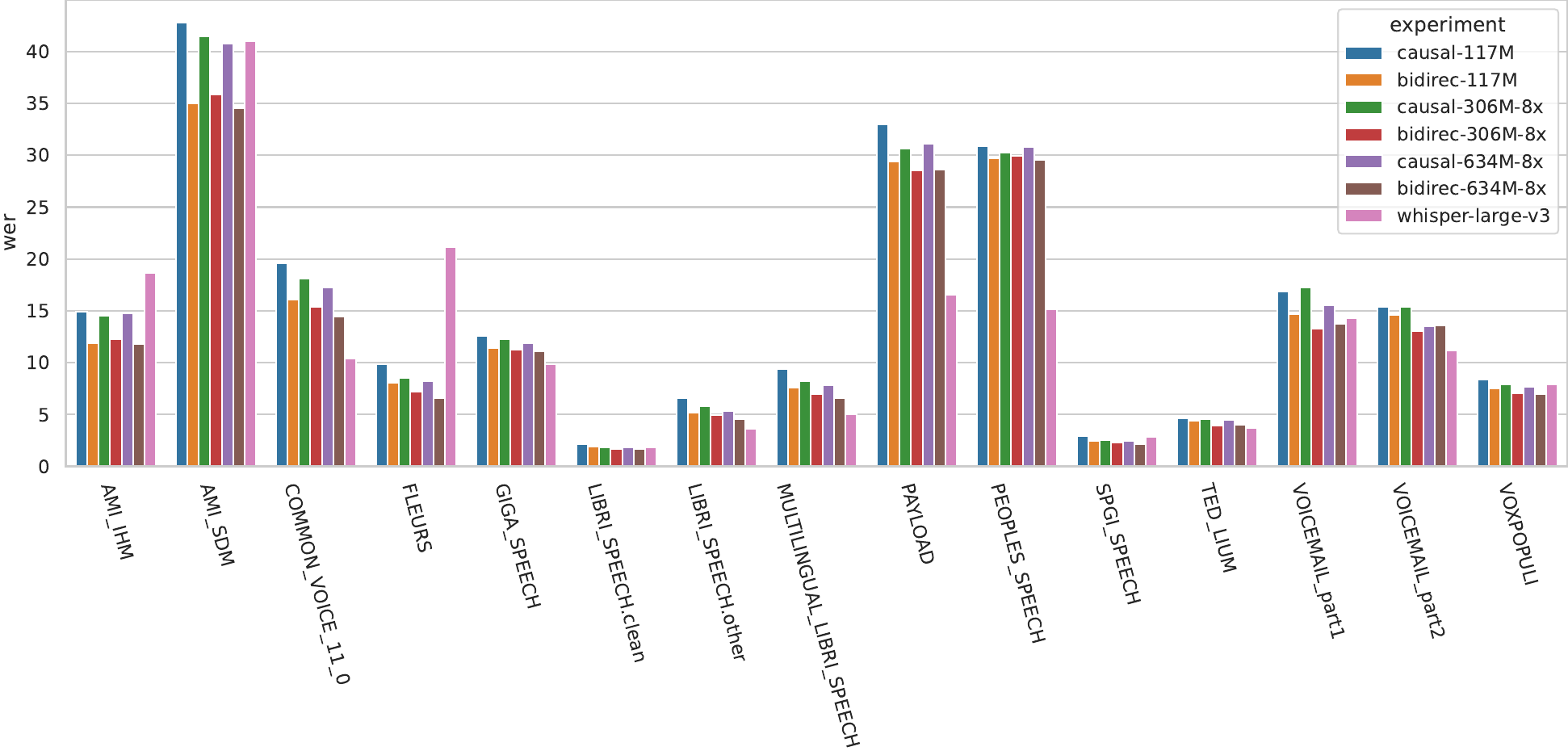}
  \caption{Word error rates of \model models compared to Whisper on English test sets.}
  \label{fig:wer}
\end{figure*}

In this work we investigate the performance of Transformer decoders and prefix LMs \cite{raffel2019exploring} as well as the individual utility of public English ASR datasets by combining them into a large 93K hour paired speech-text corpus. Unlike OWSM, we train decoder-only models solely using cross-entropy loss. We train models varying over a wide range of hyperparameters such as 1) model size, 2) bidirectionality over audio frames, 3) downsampling rate of audio frames, 4) audio augmentation and, 5) the datasets included in the training set. In addition, we also evaluate the performance of the trained models at low audio bitrates using recent neural codecs such as DAC \cite{dac}.

We find that our best \textbf{D}ecoder-\textbf{O}nly \textbf{T}ransformer for \textbf{A}SR (DOTA) model outperforms Whisper large-v3 on 7 out of 15 test sets (Figure \ref{fig:wer}), and OWSM medium-v3.1 on nearly all test sets, while having fewer than half as many parameters as Whisper. Additionally, our best \model model uses twice the audio frame downsampling rate versus OWSM, making it faster.

We have open-sourced our codebase and trained models at \url{https://github.com/ag1988/mel-asr}.



\begin{table*}[h]
  \centering
  \caption{
    Configurations of our \model models in Table \ref{tab:test-wer}. Feed-forward dimension is $4\times$ the model dimension and audio inputs are 30sec long. Audio frame rate for the model is the reciprocal of time shift.}\label{tab:model-sizes}  
  \resizebox{0.9\textwidth}{!}{%
    \begin{tabular}{@{}lcccccccc@{}}
      \toprule
    model & params (M) & layers & dimension & attention heads & embedding dim  & time shift (ms) & vocab size \\
      \midrule
      \model 117M & 117 & 16 & 768 & 12 & 128 & 40  & 30522\\
      \model 306M & 306 & 24 & 1024 & 16 & 128 & 80 & 30522\\ 
      \model 634M-8x & 634 & 32 & 1280 & 20 & 128 & 80 & 30522 \\
      \model 634M-12x & 634 & 32 & 1280 & 20 & 128 & 120 & 30522 \\
      \midrule
      OWSM medium v3.1 & 1020 & 18 & 1024 & 16 & 1024 & 40 & 50K\\ 
      Whisper medium & 769 & 24 & 1024 & 16 & 1024 & 20 & 51865\\ 
      Whisper large v3 & 1550 & 32 & 1280 & 20 & 1280 & 20 & 51866\\
       \bottomrule
    \end{tabular}%
  }
\end{table*}

\section{Method}\label{sec:method}


\subsection{Data Preparation} 

To create a large-scale supervised ASR corpus we downloaded all major public English ASR datasets and organized them into a common format. Audio was resampled to 16kHz and stored as single-channel signed 16-bit integers in HDF5 format for memory-mapped access. Our training data consists of MultilingualLibriSpeech (English) \cite{mls}, PeoplesSpeech \cite{peoplesspeech}, GigaSpeech \cite{gigaspeech}, SPGISpeech \cite{spgispeech}, CommonVoice 11.0 \cite{commonvoice}, LibriSpeech \cite{librispeech}, Fisher \cite{fisher}, TedLium 3 \cite{tedlium}, AMI \cite{ami}, FLEURS (English) \cite{fleurs}, VoxPopuli (English) \cite{voxpopuli}, LJ Speech \cite{ljspeech},  VoiceMail \cite{voicemail}, VCTK \cite{vctk}. This resulted in 93K hours of speech-text pairs (top row of Table \ref{tab:train-wer}). \\

\noindent\textbf{Text Processing}\ \ We normalized the transcripts using Whisper's \texttt{EnglishTextNormalizer} module which converts text to lower case, removes punctuation and applies several other case-by-case transformations (see \cite{whisper} for further details). We further remove newline character \texttt{`$\backslash$n'} and insert space between consecutive digits (e.g. \texttt{21} $\mapsto$ \texttt{2 1}). We then tokenize the text using \texttt{bert-base-uncased} tokenizer \cite{devlin2018bert}. Text inputs to the model are truncated to 146 tokens.

\subsection{Audio Processing}\label{sec:audio}

The audio input to the model is a 30sec waveform consisting of 480K floats. Shorter instances are 0-padded to this length and longer waveforms were truncated. Log-mel spectograms were computed using window size of 25ms and hop size 10ms using 80 mel bins. This resulted in 3000 audio frames per instance. For efficiency, we stacked every 4 frames into a single frame in the 117M model resulting in 750 audio frames. In the wider 306M and 634M models, we stacked every 8 (or 12) frames into a single frame further reducing the number of frames to 375 (or 250) (Table \ref{tab:model-sizes}).\\

\noindent\textbf{Audio Augmentation}\ \ To make the model more robust to out of distribution instances we applied each of the following augmentations to the audio waveforms during training
\begin{itemize}[leftmargin=*,topsep=0pt,itemsep=4pt,parsep=0pt]
    \item with probability (w.p.) 1e-3 we applied speed augmentation with a factor sampled uniformly from [0.9,1.1].
    \item w.p. 0.2 tempo augmentation with a factor sampled uniformly from [0.9,1.1].
    \item single-pole low pass filter w.p. 1e-3
    \item reverberation w.p. 1e-3
    \item given a partially formed training instance, we randomly concatenated the next sample from the same dataset to it w.p. $p=0.25$ or else 0-padded it w.p. $1-p$, applying this process repeatedly until the length of the partially formed instance was at least 30sec, allowing us to explore the full 30sec input limit and preventing the model from overfitting to instances mostly containing trailing silence.
\end{itemize}

\subsection{Model} 

In this work, we are interested in exploring the limits of Transformer decoder-only architectures inspired by their success as language models \cite{mixtral}. Our models comprise a single decoder stack with a causal attention mask. The input to the model is a sequence of audio frames followed by text token embeddings. Each of these vectors are mapped to the model dimension using linear projections. Each frame attends to itself and the frames on its left. However, for the prefix LMs (denoted by ``bidirec'') we allow audio frames to attend to the audio frames on their right as well. For simplicity, we used sinusoidal positional embeddings.

\begin{table*}[t]
\centering
\caption{ASR performance (WER) on English test sets. \emph{Notation}: ``bidirec'': prefix LM where causal attention mask is not used across audio frames allowing past audio frames to view future audio frames, ``no-people'': PeoplesSpeech was excluded from training set, ``no-mls'': MultilingualLibriSpeech was excluded from training set, ``eval DAC 6kbps'': test set audio was compressed via 16kHz 6kbps DAC. Whisper v3 uses 1M hours of weakly-labelled data and 4M hours of pseudo-labelled audio, OWSM uses 180K hours of public multilingual data and \model models use 93K hours of public English data. OWSM results are reported directly from \cite{owsm2} whereas the Whisper results are computed using the official repository. \model uses greedy decoding.}\label{tab:test-wer}
\resizebox{\textwidth}{!}{
\begin{tabular}{l|rrrrrrrrrrrrrrr}
\toprule
\setlength{\tabcolsep}{2pt}\renewcommand{\arraystretch}{1.2}
Model
& \multicolumn{1}{c}{\rotatebox[origin=rc]{270}{MultilingualLibriSpeech.en}}
& \multicolumn{1}{c}{\rotatebox[origin=rc]{270}{PeoplesSpeech}}
& \multicolumn{1}{c}{\rotatebox[origin=rc]{270}{GigaSpeech}}
& \multicolumn{1}{c}{\rotatebox[origin=rc]{270}{SPGISpeech}}
& \multicolumn{1}{c}{\rotatebox[origin=rc]{270}{CommonVoice 11.0.en}}
& \multicolumn{1}{c}{\rotatebox[origin=rc]{270}{LibriSpeech.test-clean}}
& \multicolumn{1}{c}{\rotatebox[origin=rc]{270}{LibriSpeech.test-other}}
& \multicolumn{1}{c}{\rotatebox[origin=rc]{270}{TED-LIUM3}}
& \multicolumn{1}{c}{\rotatebox[origin=rc]{270}{AMI-IHM}}
& \multicolumn{1}{c}{\rotatebox[origin=rc]{270}{AMI-SDM}}
& \multicolumn{1}{c}{\rotatebox[origin=rc]{270}{VoxPopuli.en}}
& \multicolumn{1}{c}{\rotatebox[origin=rc]{270}{Fleurs.en}}
& \multicolumn{1}{c}{\rotatebox[origin=rc]{270}{Voicemail.part1}}
& \multicolumn{1}{c}{\rotatebox[origin=rc]{270}{Voicemail.part2}}
& \multicolumn{1}{c}{\rotatebox[origin=rc]{270}{Payload (internal)}}
\\ \midrule
\model causal-117M &  9.4 & 30.9 & 12.6 & 2.9 & 19.6 & 2.1 & 6.6 & 4.6 & 14.9 & 42.8 & 8.4 & 9.8 & 16.8 & 15.4 & 33.0 \\
\model causal-117M-no-augmentation &  9.2 & 30.8 & 12.4 & 3.0 & 19.1 & 2.3 & 6.5 & 4.6 & 12.9 & 39.7 & 8.2 & 9.9 & 18.6 & 22.1 & 33.9 \\
\model causal-117M-no-people &  9.2 & 31.4 & 12.6 & 2.9 & 19.2 & 2.1 & 6.2 & 4.9 & 19.1 & 46.9 & 9.0 & 9.2 & 16.9 & 15.9 & 33.0  \\
\model causal-117M-no-people-no-mls &  12.5 & 30.9 & 13.0 & 3.0 & 21.1 & 3.0 & 9.0 & 5.4 & 18.6 & 45.1 & 10.2 & 10.5 & 18.1 & 15.9  & 36.0\\
\model bidirec-117M &  7.6 & 29.7 & 11.4 & 2.4 & 16.1 & 1.9 & 5.2 & 4.4 & 11.9 & 35.0 & 7.5 & 8.0 & 14.7 & 14.6 & 29.4 \\
\model bidirec-117M (eval DAC 6kbps) &  8.4 & 32.4 & 12.2 & 3.0 & 18.1 & 2.1 & 6.1 & 4.6 & 16.0 & 50.7 & 7.8 & 8.2 & 15.3 & 14.4 & 31.4 \\
\model causal-306M-8x &  8.2 & 30.3 & 12.3 & 2.5 & 18.1 & 1.8 & 5.8 & 4.6 & 14.5 & 41.5 & 7.9 & 8.5 & 17.3 & 15.4 & 30.6 \\
\model bidirec-306M-8x &  7.0 & 30.0 & 11.3 & 2.3 & 15.4 & 1.7 & 4.9 & 4.0 & 12.3 & 35.8 & 7.1 & 7.2 & 13.2 & 13.0 & 28.6 \\
\model causal-634M-8x &  7.8 & 30.8 & 11.9 & 2.4 & 17.2 & 1.9 & 5.3 & 4.5 & 14.7 & 40.8 & 7.7 & 8.2 & 15.5 & 13.5 & 31.1  \\
\model bidirec-634M-8x &  6.6 & 29.6 & 11.1 & \textbf{2.1} & 14.5 & \textbf{1.7} & 4.6 & 4.0 & \textbf{11.8} & \textbf{34.5} & \textbf{6.9} & 6.6 & 13.7 & 13.6 & 28.6  \\
\model bidirec-634M-8x (eval DAC 6kbps) &  7.2 & 31.3 & 11.5 & 2.5 & 16.2 & 1.9 & 5.1 & 4.1 & 14.7 & 49.0 & 7.1 & 7.2 & 14.0 & 12.8 & 30.0 \\
\model bidirec-634M-12x &  6.8 & 29.8 & 11.4 & 2.2 & 15.3 & \textbf{1.7} & 4.7 & 4.0 & 12.6 & 35.9 & 7.2 & 7.1 & \textbf{12.7} & 12.7 & 32.5  \\
\midrule
OWSM v3.1 medium (1.02B) & 7.1  &  &  &  & 12.6 & 2.4 & 5.0 & 5.1 &  &  & 8.4 & 9.0 &  &  & 
\\ 
Whisper large v2 (1550M) &  6.8 & 18.0 & 10.5 & 3.9 & 10.6 & 2.6 & 4.9 & 4.1 & 19.4 & 41.8 & 7.6 & 13.8 & 15.6 & 12.1 & 19.3
\\
Whisper large v3 (1550M) &  \textbf{5.0} & \textbf{15.2} & \textbf{9.9} & 2.8 & \textbf{10.4} & 1.8 & \textbf{3.6} & 3.7 & 18.7 & 41.0 & 7.9 & 21.2 & 14.3 & \textbf{11.2} & \textbf{16.6}
\\
Whisper large v3 (eval DAC 6kbps) &  5.2 & 16.0 & 10.0 & 2.9 & 11.1 & 1.9 & 4.1 & \textbf{3.6} & 16.2 & 38.0 & 6.8 & \textbf{4.1} & 15.0 & 12.1 & 17.3  \\
\bottomrule
\end{tabular}
}
\end{table*}

\begin{table*}[t]
\centering
\caption{WER computed over at most 24K samples chosen randomly from each training set.}\label{tab:train-wer}
\resizebox{1.01\textwidth}{!}{
\begin{tabular}{l|rrrrrrrrrrrrrrrrrrrrr}
\toprule
& \multicolumn{1}{c}{\rotatebox[origin=rc]{270}{MultilingualLibriSpeech.en}}
& \multicolumn{1}{c}{\rotatebox[origin=rc]{270}{PeoplesSpeech.clean}}
& \multicolumn{1}{c}{\rotatebox[origin=rc]{270}{PeoplesSpeech.clean.sa}}
& \multicolumn{1}{c}{\rotatebox[origin=rc]{270}{PeoplesSpeech.dirty}}
& \multicolumn{1}{c}{\rotatebox[origin=rc]{270}{PeoplesSpeech.dirty.sa}}
& \multicolumn{1}{c}{\rotatebox[origin=rc]{270}{GigaSpeech}}
& \multicolumn{1}{c}{\rotatebox[origin=rc]{270}{SPGISpeech}}
& \multicolumn{1}{c}{\rotatebox[origin=rc]{270}{CommonVoice 11.0.en}}
& \multicolumn{1}{c}{\rotatebox[origin=rc]{270}{Fisher}}
& \multicolumn{1}{c}{\rotatebox[origin=rc]{270}{LibriSpeech.clean.100}}
& \multicolumn{1}{c}{\rotatebox[origin=rc]{270}{LibriSpeech.clean.360}}
& \multicolumn{1}{c}{\rotatebox[origin=rc]{270}{LibriSpeech.other.500}}
& \multicolumn{1}{c}{\rotatebox[origin=rc]{270}{TED-LIUM3}}
& \multicolumn{1}{c}{\rotatebox[origin=rc]{270}{AMI-IHM}}
& \multicolumn{1}{c}{\rotatebox[origin=rc]{270}{AMI-SDM}}
& \multicolumn{1}{c}{\rotatebox[origin=rc]{270}{VoxPopuli.en}}
& \multicolumn{1}{c}{\rotatebox[origin=rc]{270}{Fleurs.en\_us}}
& \multicolumn{1}{c}{\rotatebox[origin=rc]{270}{Voicemail.part1}}
& \multicolumn{1}{c}{\rotatebox[origin=rc]{270}{Voicemail.part2}}
& \multicolumn{1}{c}{\rotatebox[origin=rc]{270}{LJSpeech}}
& \multicolumn{1}{c}{\rotatebox[origin=rc]{270}{VCTK}}
\\ 
\midrule
dataset size (hours) &  41K & 6K & 1K & 22K & 2K & 10K & 5K & 1.5K & 2K & 100 & 360 & 500 & 453 & 78 & 77 & 522 & 7 & 15 & 15 & 23 & 82
\\ \midrule
\model causal-117M &  2.6 & 5.9 & 5.6 & 17.9 & 17.7 & 3.5 & 2.4 & 5.3 & 8.6 & 1.5 & 1.6 & 2.2 & 3.8 & 8.1 & 32.3 & 5.5 & 5.0 & 7.0 & 8.1 & 0.9 & 0.5  \\
\model causal-117M-no-augment &  2.7 & 5.8 & 5.5 & 17.7 & 17.4 & 4.0 & 2.6 & 7.6 & 9.6 & 1.6 & 1.7 & 2.3 & 4.5 & 9.3 & 33.0 & 5.9 & 6.0 & 8.9 & 10.0 & 1.1 & 0.7  \\
\model causal-117M-no-people &  2.3 & 10.1 & 9.3 & 24.7 & 24.0 & 3.0 & 2.1 & 3.4 & 7.3 & 1.2 & 1.3 & 1.8 & 3.3 & 9.6 & 34.7 & 5.2 & 3.9 & 5.4 & 6.6 & 1.7 & 0.3  \\
\model causal-117M-no-people/mls &  4.9 & 10.4 & 9.6 & 25.1 & 24.4 & 2.1 & 1.6 & 1.3 & 3.9 & 1.4 & 1.5 & 2.1 & 1.9 & 6.3 & 24.6 & 3.5 & 2.1 & 2.6 & 3.7 & 1.7 & 0.1  \\
\model bidirec-117M &  2.4 & 5.6 & 5.2 & 17.2 & 16.7 & 3.0 & 2.1 & 4.7 & 7.9 & 1.3 & 1.4 & 1.8 & 3.3 & 5.8 & 24.0 & 5.2 & 4.7 & 6.2 & 7.3 & 0.7 & 0.4  \\
\model causal-306M-8x &  2.2 & 5.6 & 5.2 & 16.9 & 16.5 & 2.7 & 1.9 & 2.8 & 6.6 & 1.1 & 1.1 & 1.6 & 2.7 & 5.0 & 24.9 & 4.5 & 3.4 & 4.6 & 5.6 & 0.6 & 0.2  \\
\model bidirec-306M-8x &  2.1 & 5.4 & 5.0 & 16.4 & 16.0 & 2.5 & 1.8 & 3.1 & 6.9 & 1.1 & 1.1 & 1.5 & 2.7 & 4.1 & 19.9 & 4.4 & 3.4 & 4.4 & 5.6 & 0.5 & 0.2  \\
\model causal-634M-8x &  2.0 & 5.2 & 4.9 & 16.3 & 15.9 & 2.1 & 1.6 & 1.6 & 5.0 & 0.9 & 0.9 & 1.3 & 2.0 & 3.8 & 20.5 & 3.6 & 2.5 & 3.0 & 4.5 & 0.4 & 0.1  \\
\model bidirec-634M-8x &  2.0 & 5.1 & 4.6 & 15.8 & 15.3 & 2.1 & 1.6 & 1.7 & 5.1 & 0.9 & 1.0 & 1.2 & 1.9 & 2.1 & 13.9 & 3.5 & 2.5 & 2.9 & 4.2 & 0.4 & 0.1  \\
\model bidirec-634M-12x &  1.9 & 5.2 & 4.7 & 16.1 & 15.6 & 2.2 & 1.6 & 1.9 & 5.4 & 0.9 & 1.0 & 1.2 & 2.2 & 2.6 & 15.8 & 3.7 & 2.5 & 3.1 & 4.6 & 0.3 & 0.1 \\
\midrule
Whisper large v3 &  2.3 & 9.6 & 9.5 & 23.8 & 23.1 & 4.3 & 2.9 & 6.3 & 15.1 & 1.6 & 1.6 & 2.0 & 5.1 & 20.0 & 41.4 & 7.4 & 4.6 & 9.9 & 10.8 & 1.7 & 1.2 \\
\bottomrule
\end{tabular}
}
\end{table*}

\subsection{Training} 

Our models are trained for 1M steps with a batch size of 128 using AdamW optimizer with moment scales $\beta_1=0.9$, $\beta_2=0.99$. The peak learning rate was 2e-4 linearly warmed up over 1K steps and decayed to 0 using cosine schedule. Weight decay of 0.1 was used but was not applied to biases and layer norm parameters. Training was performed in bfloat16 precision and the gradients were clipped to norm 1.0. We did not use early stopping and used the final checkpoint in all evaluations. The largest \model models, bidirec-634M-12x and bidirec-634M-8x, were trained on 8 40GiB A100's for 85 and 120 hours respectively. During training, all datasets except MultilingualLibriSpeech, PeoplesSpeech, GigaSpeech, SPGISpeech and LibriSpeech, were upsampled by $2\times$ to account for the large differences in the dataset sizes.

We experimented with various model and training configurations as shown in Table \ref{tab:test-wer} with the corresponding model sizes in Table \ref{tab:model-sizes}.

\section{Speech Recognition Performance}\label{sec:results}

To evaluate the performance of the trained models, we performed inference on the test and training sets using greedy decoding and computed the word error rates (WER) as shown in Table \ref{tab:test-wer} and Table \ref{tab:train-wer}. Due to the large sizes of training sets, we randomly sampled 24K instances for the training sets of size larger than this amount. During evaluation we did not perform any augmentations listed in \S\ref{sec:audio} and for all evaluated models removed instances where the reference transcript contained more than 145 tokens. This removal did not affect the evaluation on the test sets.

As shown in Table \ref{tab:test-wer}, our \model models are competitive with the best performing ASR models such as Whisper, while having significantly fewer parameters. Concretely, our best model \model bidirec-634M-8x outperforms Whisper large-v3 on 7 out of 15 test sets despite the latter being trained on 10$\times$ more data. Moreover, our bidirec-634M-8x model outperforms OWSM on all the test sets on which its authors reported results with the exception of CommonVoice (En).\\

\noindent\textbf{Bidirectionality}\ \ Comparing the performance of the ``causal'' \model models vs the corresponding prefix LMs (``bidirec'') we find that bidirectionality over audio frames is critical to high performance across model scales. Surprisingly, even the smallest prefix LM  \model bidirec-117M outperforms a significantly larger causal model \model causal-634M-8x.\\

\noindent\textbf{Dataset ablation}\ \ Our baseline causal model causal-117M is trained on 93K hours of paired data. To determine the utility of the largest components of our training data viz. MutlilingualLibriSpeech (MLS) and PeoplesSpeech of sizes 41K hours and 29K hours respectively we trained two analogous variants after ablating these datasets. As shown in Table \ref{tab:test-wer}, excluding PeoplesSpeech (causal-117M-no-people) did not result in significant degradation on most test sets with the exception of AMI test sets containing meeting recordings. However, further excluding MLS (causal-117M-no-people-no-mls) resulted in a significant degradation w.r.t. the baseline model. \\

\noindent\textbf{Augmentation}\ \ We trained a variant causal-117M-no-augmentation of our baseline causal model without using the augmentations described in \S\ref{sec:audio} and found their performance to be similar on the test sets.\\

\noindent\textbf{Performance on training sets}\ \ The performance on the training sets is summarized in Table \ref{tab:train-wer}. As our models are trained on these sets, the errors rates are low on most training sets with the exception of the ``dirty'' splits of PeoplesSpeech. However, compared to our \model models, Whisper reports significantly higher error rates with performance gaps as large as 10 on Fisher (telephony) and 25 on AMI-SDM (microphone).\\

\noindent\textbf{Audio at low bitrate}\ \ We also directly evaluated our trained models after encoding the audio via recent neural audio codecs such as DAC \cite{dac} which have reported significantly better performance at lower bitrates compared to commonly used codecs such as Opus. We used the 16kHz 6kbps version of DAC and include the results in Table \ref{tab:test-wer}. For our \model models, we observe significant degradation on test sets such as AMI (meetings) highlighting the importance of exposing the codec to diverse modalities during training. However, we observe no such degradations in case of Whisper pointing to the possibility of Whisper being exposed to low bitrate audio during training which was not the case for our \model models. This is an interesting development as it is significantly easier to scale up dataset sizes at lower bitrates due to the smaller download size and disk footprint. In the future, we plan on mixing in lower bitrate audio while training \model models.


\comment{
\section{Acknowledgments}
}

\bibliographystyle{acl_natbib}

\bibliography{all}


\end{document}